\documentclass[sigconf]{acmart}

\AtBeginDocument{%
  \providecommand\BibTeX{{%
    \normalfont B\kern-0.5em{\scshape i\kern-0.25em b}\kern-0.8em\TeX}}}

\copyrightyear{2020}
\acmYear{2020}
\setcopyright{acmcopyright}\acmConference[HuMA'20]{1st International Workshop on Human-centric Multimedia Analysis}{October 12, 2020}{Seattle, WA, USA}
\acmBooktitle{1st International Workshop on Human-centric Multimedia Analysis (HuMA'20), October 12, 2020, Seattle, WA, USA}
\acmPrice{15.00}
\acmDOI{10.1145/3422852.3423481}
\acmISBN{978-1-4503-8151-2/20/10}

\begin{document}
\title{Human-Object Interaction Detection: \\A Quick Survey and Examination of Methods}

\author{Trevor Bergstrom, Humphrey Shi}
\email{{bergsttr,hshi3}@cs.uoregon.edu}
\affiliation{%
  \institution{University of Oregon}
  \streetaddress{1477 E 13th Ave}
  \postcode{97403}
}


\renewcommand{\shortauthors}{Bergstrom et al.}

\begin{abstract}
  Human-object interaction detection is a relatively new task in the world of computer vision and visual semantic information extraction. With the goal of machines identifying interactions that humans perform on objects, there are many real-world use cases for the research in this field. To our knowledge, this is the first general survey of the state-of-the-art and milestone works in this field. We provide a basic survey of the developments in the field of human-object interaction detection. Many works in this field use multi-stream convolutional neural network architectures, which combine features from multiple sources in the input image. Most commonly these are the humans and objects in question, as well as a spatial quality of the two. As far as we are aware, there have not been in-depth studies performed that look into the performance of each component individually. In order to provide insight to future researchers, we perform an individualized study that examines the performance of each component of a multi-stream convolutional neural network architectures for human-object interaction detection. Specifically we examine the HORCNN architecture as it is a foundational work in the field. In addition, we provide an in-depth look at the HICO-DET dataset, a popular benchmark in the field of human-object interaction detection. 
  Code and papers can be found at {\color{red} \href{https://github.com/SHI-Labs/Human-Object-Interaction-Detection}{https://github.com/SHI-Labs/Human-Object-Interaction-Detection}}.
\end{abstract}



\maketitle

\section{Introduction}
Achieving the goal of true machine intelligence requires an agent that can observe and understand its environment just as humans are able to. There has been a significant amount of excitement and progress around machine learning and its ability to solve problems related to emulating human understanding of our natural and social environments. The field of computer vision, in particular, has recently exploded with the advent of deep learning techniques that can perform well on complex object detection problems. However, simply identifying objects in an image is not what should be considered true machine intelligence. Striving towards the idea of more intelligent machines, researchers have created models and systems that can extract richer semantic information from images and videos. As humans, we are able to recognize relationships between objects in an image. These relationships can help an intelligent machine interpret the underlying meaning of the image or scene, and therefore, take one step closer to understanding the world around us. 

We choose to divide the main track of AI computer vision research into two tracks, visual perception tasks, and visual understanding tasks. Visual perception tasks focus on identifying parts or features of an image. Object detection and image segmentation are examples of well known domains that fit in the purview of visual perception. Given a scene or image, our goal is to make some quick observations that are easy to see. For example, we should be able to determine the possible objects present in an image, or whether or not a section of the image is part of an object or the background. These properties must to be learned or hand designed, but generally are not more complex than reinforcing specific feature combinations that make up a human, while a different combination of features make up a cat. Visual understanding tasks on the other hand, require far more complex analysis of a scene. These tasks focus on the less visual and at as easily recognized features of an image. Visual understanding tasks include domains such as visual relationship detection and activity recognition, as well as our focus of human-object interaction detection. In many of these tasks, visual perceptions such as object detection, are prerequisites to obtain before moving on to identifying the finer grained features needed to complete the task. For human-object interaction detection, a model must detect the possible objects in the image, then make a decision on whether an interaction is occurring and if so what that interaction is.

Human-object interaction detection is closely related to other computer vision research areas such as visual relationship detection and activity recognition. However, the foundational works and datasets differ from those used in the aforementioned tasks. We feel that there is a need for a brief and general overview for future researchers to easily obtain the baseline knowledge to create contributions in this field. One of our main contributions is a quick survey of deep learning based methods for solving human-object interaction detection, as well as a look at some of the datasets and metrics used for evaluating models. Commonly in this field, classifying the interactions is done through the use of a multi-stream convolutional neural network. We present a detailed examination of the performance of each individual stream component. This investigation can provide useful information on how to develop future models, using this multi-stream method. Using the components from HORCNN \cite{chao2018learning}, we conduct tests on their ability to correctly classify human-object interactions.

We also provide a careful analysis of the HICO-DET dataset \cite{chao2018learning}. This dataset is commonly used as a baseline for human-object interaction detection. This dataset contains a large and complex set of interactions on various objects. We provide a meaningful analysis of the various components, and suggestions for improving this state-of-the-art dataset. 

\section{Human-Object Interaction Detection Domain Survey}
When humans seek to interpret their environment, they do so by observing other humans and how they interact with one another or objects. Object to object interactions, for the most part, deal with simple spatial or descriptive interactions. Humans can provide a much richer set of interactions with objects, as there are visual and non-visual ways a human can interact with their natural environment. This work will focus primarily on the task of human-object interaction detection. The goal of human-object interaction detection, is to correctly identify humans, objects, and the actions that are occurring between them, if any, in an image \cite{chao2015hico}. The first step in discovering an human-object interaction from an image is to detect objects. Object proposals recovered from the image should contain at least one human for an human-object interaction to be present. Using these human and object proposals, a model for solving this problem must then correctly identify a human-object interaction between the humans present and any of the objects in the image. 

We have classified the methods of solving human-object interaction detection problems into the two classes: multi-stream architectures and graph networks. Multi-stream architectures produce promising results and are easily augmented with supplemental information detection methods such as pose and gaze. Graph neural networks intuitively connect objects in the image in a graphical form of nodes and connected images, that represent the relationships between objects in the image. This section will provide further insight into how each of these approaches identifies human-object interactions, as well as their strengths and drawbacks. 

\subsection{Multi-Stream Approaches}
\begin{figure}
\centering
  \includegraphics[width=0.8\columnwidth]{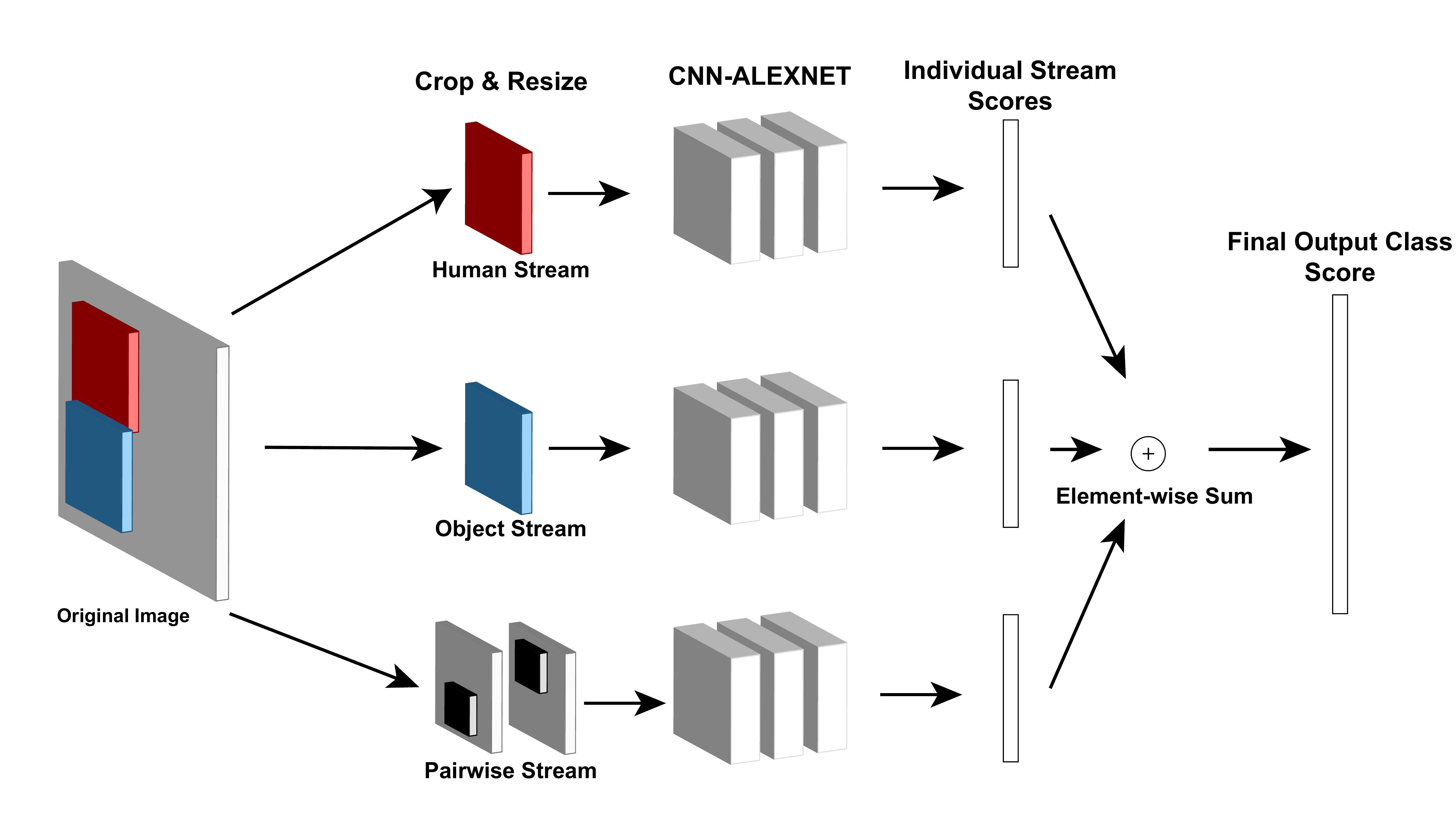}
  \caption{Diagram of the HORCNN architecture. }~\label{fig:figure1}
\end{figure}

Multi-stream convolutional neural networks were first proposed for the task of human-object interaction detection by Chao et al. as Human-Object Region-based Convolutional Neural Networks (HORCNN) \cite{chao2018learning}. HORCNN includes three "streams", based around CNN architectures, to extract features from different sources in the image. Using object proposals from the RCNN \cite{girshick2014rich} object detector, the human and object streams extract appearance queues from the image. The human stream can interpret human pose at an elementary level. For example, a person riding a bike is most likely to be in a sitting pose rather than standing. Similarly, the object stream can interpret the appearance of the object involved in the interaction. Again, using the riding-bike example, a bicycle being ridden has a higher probability of being occluded by the person in the image. The final stream in HORCNN extracts spatial information between the human and object. This may be one of the more obvious queues when inferring human-object interactions. Reusing the riding-bike example, a human riding a bike is more likely to be located on top of the bike rather than to either side if they were instead standing-next-to-bike. Both the human and object streams are based on CaffeNet \cite{jia2014caffe} implementations, pre-trained on ImageNet. At the end of each stream, classification for the possible interaction classes is performed. Finally, an element-wise sum is taken for their feature vectors for final classification scores, therefore each stream has equal an equal weight in the final classification score. Due to the multi-tasking nature of humans, HOI detection should be considered a multi-label classification problem, as a person can be performing more than one interaction on an object at a time. The individual streams and network architecture of HORCNN can be seen in Figure \ref{fig:figure1}.

Building on this method of multi-stream approach, Gkioxari et al. \cite{gkioxari2018detecting} use a similar architecture for detecting human-object interactions, in InteractNet. They use three branches based on Faster R-CNN feature extraction backbone: an object detection branch, a human-centric branch, and an interaction branch. The object detection branch is identical to Faster R-CNN \cite{ren2015faster}, performing bounding box regression, and computing a classification score for the detected  humans and objects in the image. The human centric branch performs two tasks, action classification, and target object localization. Similarly to HORCNN, human appearance features are used to compute an action classification score or the probability that the human in question is performing a specific action. Target localization again uses human appearance features, and action score to model the probability density of the target object's location. The final branch of interaction recognition combines the features detected for the human centric branch with appearance features from the target object. The score is computed by preforming sigmoid activation on the outputs from human action and target action classification. The final score is a product of the human and object detection scores, the target localization score, and the action classification score.

Another implementation of the multi-stream architecture is presented by Gao et al. Instance Centric Attention Network (iCAN) \cite{gao2018ican}. They propose using an attention-based mechanism for their architecture streams. As seen in HORCNN, the three streams used are a human, object, and spatial configuration stream, and generating proposals from the Faster R-CNN detector. The difference from HORCNN is the use of the proposed instance centric attention network, replacing the conventional CNN architectures. Unlike extracting object appearance and human appearance as individual queues, iCAN aims to extract contextual features from both the human and object instances in the image. iCAN begins by extracting the appearance features from the localized object to dynamically generate an attention map on that object instance. This is accomplished by embedding the appearance features and convolutional feature maps, measuring similarity using a dot product operation. The attention map is generated using a softmax function. A contextual feature is extracted from the attention map through the weighted average of convolutional features. The iCAN module outputs a concatenation of the instance level appearance features and the contextual appearance features. Scores for each action are computed similarly to InteractNet, combining the detection confidence, action, and target location probability scores. 

\begin{figure}
\centering
  \includegraphics[width=0.8\columnwidth]{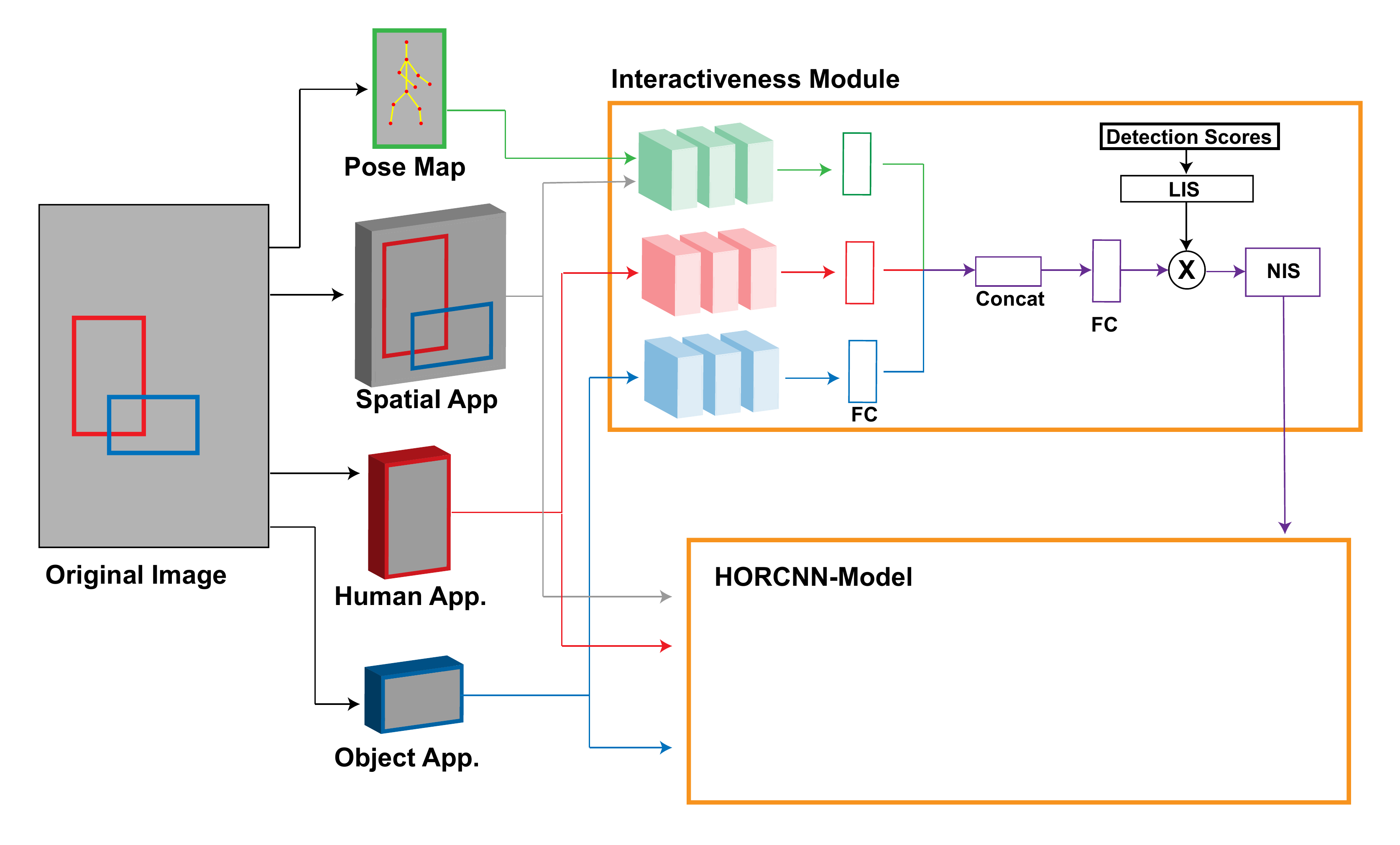}
  \caption{Diagram of the Transferable Interactiveness Network architecture. }~\label{fig:figure3}
\end{figure}

\subsubsection{Fine-Grained Information Retrieval}
It can be seen from the iCAN implementation that more information than appearance and spatial relations benefit the goal of HOI detection. There has been considerable research into using finer-grained contextual information extracted from the detected human to enhance HOI models. Yao and Fei-Fei \cite{yao2010modeling} very early on proposed the idea that pose information and object can provide mutual context to each other, showing that a detection of one can provide informative cues towards the detection of the other. They estimate human pose to help detect target objects for the human's interaction in the image using a random field model that learns the connectivity patterns between human body parts and objects. Fang et al. \cite{fang2018pairwise} propose another model that uses the individual body part attention specifically for use on human-object interaction detection datasets. The authors note that just using individual body parts attention does not capture the correlation between different body parts used in a specific interaction. Therefore, they propose generating attention maps from pairs of body parts and select specific pairs that best fit the interaction in question.

\begin{figure}
\centering
  \includegraphics[width=0.9\columnwidth]{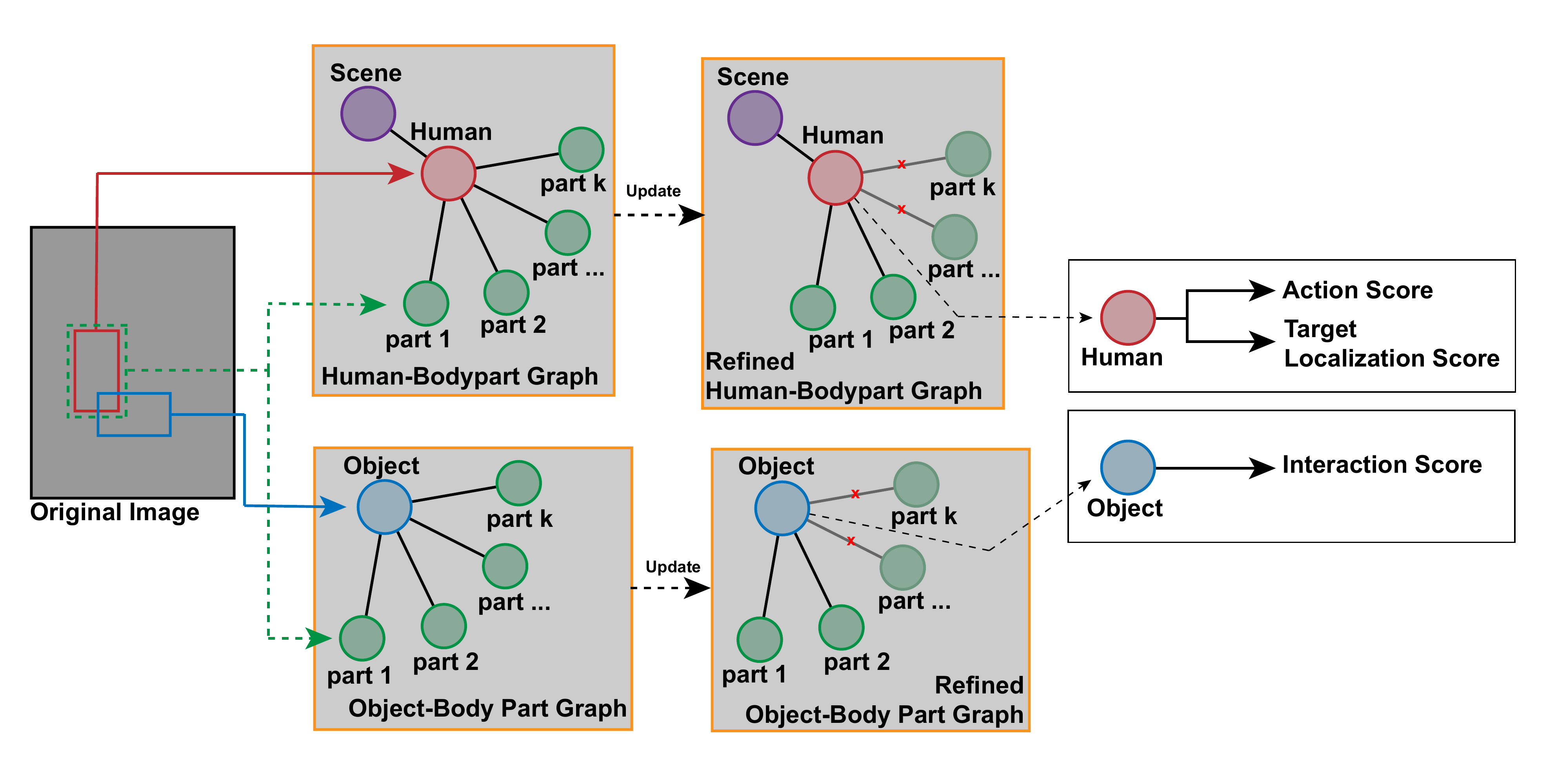}
  \caption{Diagram of the RPNN graph parsing process. }~\label{fig:figure2}
\end{figure}

Many works have proposed using human pose estimation to aid in detection results, some of the first being Gupta et al. \cite{gupta2018no} and Li et al. \cite{li2018transferable}. Li et al. propose an add-on module to existing human-object interaction detection models, using pose information as supplemental information in their spatial information stream. The Transferable Interactiveness Network (TIN) module uses a three stream, convolutional feature extraction architecture similar to HORCNN, combining a human pose map with their spatial configuration stream. Their network works on the idea of eliminating pairs of humans and objects that are not likely to be interacting with each other. Specifically they use two functions, the low-grade suppressive (LIS) function which determines the interactivness from the detection scores, and the non-interaction suppression function which eliminates the pairs of humans and objects that are not interacting. The final outout score is incorporated into an existing model such as HORCNN as the authors used in their work. It should be noted that the interactiveness score only applies to HOIs in which the human physically interacts with an object to produce the interaction. Therefore, only these interactions can benefit from this method. Li et al. also incorporate a knowledge transfer training mechanism that influences the Interactivness Network module. This mechanism provides learned information from multiple human-object interaction datasets to produce a highly accurate inference on a testing image. The architecture of this add on module can be seen in Figure \ref{fig:figure2}.

Another model that uses pose estimation is the Pose-aware Multi-level Feature Network or PMFNet proposed by Wan et al. \cite{wan2019pose}. This approach utilizes a slightly different architecture and score fusion than previously examined in this survey. PMFNetbuilds upon the method of body part attention maps, but not constrained to pairs as in the Interactiveness Network. Additionally, spatial relations between body parts and the object in question are computed to encode fine spatial configuration information. The multi-stream architecture employs three modules, a holistic module, a zoom-in module, and a fusion module for feature fusion. Using human, object, and union (interaction area) proposals detected using Faster R-CNN \cite{ren2015faster} as an object detector, a conventional CNN architecture is used to extract appearance features. This same CNN also extracts a spatial configuration map between the human and the objects. The authors use the CPN pose estimator developed by Chen et al. \cite{chen2018cascaded}. The spatial features, appearance features, and pose estimation are fed to the holistic and zoom-in modules. The holistic module aims to capture object level and related context information, consisting of four streams: human, object, union, and spatial configuration. Each stream is responsible for embedding their respective output features. These are concatenated to create a holistic feature representation. The zoom-in module us responsible for extracting fine-grained information from the human pose spatial configuration, considered human body part-level features. This module contains three branches that extract human part level appearance features, human part level spatial configuration features, and an attention component to enhance relevant human parts to each specific interaction. These features are concatenated to result in the local feature representation. In the final fusion module, both the local features and the holistic features are used to fuse relation reasoning from both the coarse level and fine level features. The first benefit of this module is the ability to use coarse features as a contextual cue to suppress interactions that cannot exist in the current set of human and object proposals, this is denoted as an interaction affinity score. The other benefit is an ability to use both object level and part-level features to determine the relation score from fine-grained representations, denoted as the local relation score. Both the interaction affinity score and the local relation scores are fused to create a final score for the interaction given the human and object proposals. 

One method of note proposed by Xu et al. \cite{xu2019interact},  Intention Driven Human-Object Interaction Detection or iHOI, incorporates the features obtained from human gaze following. This is done through another multi-stream architecture. First, a set of visual and spatial features are extracted using established methods. As is common in human-object interaction detection, Faster-RCNN \cite{ren2015faster} is used to create human and object proposals. A pose estimation network from \cite{chen2018cascaded}, and a gaze direction detector borrowed from \cite{valenti2011combining}, are trained on other datasets and used to extract human body joint locations and gaze target location respectively. These features are combined into three separate streams in the model. An individual stream for extracting appearance features from both the human and object, a human-object pairwise stream for extracting features from the spatial configurations and appearances of the human and the object together, and finally a gaze driven context-aware branch that aims to infer the focus area of the human through body positioning and through the gaze location. These features are then combined to create a final human-object interaction prediction. However, iHOI does not improve performance of human-object interaction detection much beyond its contemporary counter parts. There has been some discussion of integrating more modern gaze following algorithms such as \cite{park2018learning}, \cite{yu2018deep}, \cite{liang2020pose}, or \cite{zhang2015appearance}. However, these approaches are considered slow, needing many network streams and extra processing to make a final prediction. 

A recent model by Zhou et al. \cite{zhou2020cascaded}, Cascaded HOI proposes a very complex multi-stream network architecture, incorporating language priors, geometric features, and visual features to achieve a high score on the V-COCO dataset. Their visual feature module includes using gaze type cues as well as pose estimation features to create a very robust prediction based on just the visual information present in the image. The geometric feature branch is strikingly similar to the spatial or pairwise streams of previous models like \cite{chao2018learning} and \cite{gkioxari2018detecting}. Another work called Parallel Point Detection and Matching (PPDM) \cite{liao2020ppdm}, use purely spatial features to predict the interaction class between humans and the objects. They also implement a novel hourglass shaped neural network backbone for their model. PPDM performs well on HICO-DET dataset.

\subsection{Graph Neural Networks}
An image with human-object interactions can be interpreted similarly to a scene graph, in which the nodes represent objects and humans while the edges connecting the nodes represent relations between them. This method is very similar to the task of scene graph generation, which is followed very closely in the human-object interaction detection task by \cite{xu2019learning}, breaking the task down into a graph. Qi et al. \cite{Qi_2018_ECCV} propose a novel model using a graph neural network based on message passing. The goal of the model, called the Graph Parsing Neural Network (GPNN), is to take a complete human-object graph of the image that includes all possible interactions between the human and the objects and remove edges that represent non-existing interactions in the image. This structure enables the model to preserve spatial relationships while detecting human-object interactions. GPNN generates the graph structure through the use of a link function. Then the message, update, and readout functions are used in belief propagation. The message function summarizes messages or information coming from other connected nodes, while the update function updates the hidden node states according to the incoming information. The final readout function generates an output label based on the hidden node states. Each function uses various neural network architectures as detailed in their paper. The probability of an HOI occurring between nodes is a product of the final output probabilities between the human and object nodes. 

Using the idea of graph neural networks, Zhou et al. \cite{zhou2019relation} provide an improvement on the GPNN \cite{Qi_2018_ECCV} model. Known as the relation parsing neural network (RPNN), this network focuses around two graphs, an object body part graph and a human body part graph. The object body part graph describes the relationships expressed in the image between body parts of a specific human and the surrounding objects in the image. The human body part graph models the relationship between the human and their body parts, similar to the task of pose estimation, to describe the actions and movements of the human as they relate to a specific interaction. The two graphs are fused using a message passing mechanism like in GPNN to convey information for a final interaction class prediction. This network models body part contexts to predict actions. RPNN performs very well on HICO-DET and V-COCO. A visual illustration on how RPNN parses the two graphs can be seen in Figure \ref{fig:figure3}. A more recent work into graph neural networks was conducted by Liang et. al \cite{liang2020visual}, earning this paper a top mAP score for the HICO-DET dataset. Like RPNN, they use a dual graph strategy with semantic information coming from the class labels and visual information to construct a final optimized scene graph of each object and human in the image. This model currently has the highest performance score on the HICO-DET dataset. Graph neural networks seem to be outperforming other methods for human-object interaction detection, there have been many recent works that exploit them as well as other information such as pose estimation, \cite{zheng2020skeleton} is a good example of this. 

\subsection{Weakly Supervised Approaches}
An interesting area of computer vision research is in the area of weakly supervised and zero-shot approaches to learning. Weak supervision entails that a learning algorithm is given very few training examples of a specific task, such as identifying objects. Zero-shot signifies that the specific example has never been seen by the algorithm. Both weak supervision and zero-shot approaches have been well documented throughout the years for more classical tasks of computer vision, such as object detection \cite{wei2018ts2c} and segmentation \cite{wei2018revisiting, qian2019weakly}, or even without the use of deep neural networks as in \cite{bilen2015weakly} and \cite{lampert2013attribute}, and using autoencoders as seen in \cite{kodirov2017semantic}. 

Specifically for human-object interaction detection, zero-shot and weakly supervised learning techniques are useful due to most datasets expressing a long-tailed distribution of image data. The long-tailed distribution describes the greater prevalence of common examples in the data than that of more uncommon examples. For example, there are many more examples of human-ride-horse than examples of human-ride-zebra, both because of the rarity of zebras and the rarity of scenarios where a human would be riding a zebra. However, the example of human-ride-zebra is not an impossible scenario, and a well generalized model should be able to identify these rare relationships just as humans can. This long-tailed distribution in datasets reflect the real-world, where we know that some interactions are rarer than other. For visual understanding tasks this process becomes more difficult as it is harder to rely on well-defined visual features such as those generated by SIFT \cite{lowe1999object} features or convolutional neural networks  \cite{zhang2017ppr}. However some distribution issues can be attributed to the dataset, as seen in the study \cite{kilickaya2020diagnosing}, exploring HICO-DET and some of the multi-stream models covered in this survey. 
An attempt at the task of zero-shot recognition and weakly supervised learning is seen by Pyere et al. in \cite{peyre2017weakly}, incorporating sematic language information from large text databases that provide probabilities for the interaction in question. One very early example of a weakly supervised approach is seen by Prest et al. in \cite{prest2011weakly} using a probabilistic type model, however it has not been tested on modern datasets such as HICO-DET.

More recent work seems to focus on improving these zero-shot interaction classes, and these improvements even help overall generalization on most datasets, this improvement can be seen in works such as \cite{bansal2018zero}, \cite{wang2020discovering}, and \cite{kato2018compositional}. Hou et al. \cite{hou2020visual} propose the Visual Compositional Learning (VCL) framework for human-object interaction detection. Their network learns shared object and verb features, breaking down verbs to relate to specific objects. This process learns shared object and verb features from across all human-object interactions. Their framework uses another multi-stream process containing three streams. Specifically their main contribution is their verb-object branch that extracts verb or interaction class features from the union of both the human and object bounding boxes. They show superior performance on the HICO-DET dataset using this method. In a closely related approach Bansal et al. \cite{bansal2020detecting} use the idea of the similarities between human actions to help guide zero-shot classes using human features from more common instances, just like the example given of human-ride-horse and human-ride-zebra. They use a word2vec model to model similarities between objects which provides a likelihood that a specific interaction can occur between the object in question and the human, based on other similar objects. Their work shows improvement on the HICO-DET dataset and for zero-shot test cases, but they detail their approach as being limited by the fact that an interaction can look entirely different on two seemingly similar objects. Another interesting recent work on improving generalization across the lesser seen interaction examples is done by Song et al. in \cite{song2020novel}. They propose using adversarial domain generalization to encourage predictions on the unseen or longer tailed examples. Specifically they focus on improving the spatial stream in a network similar to that of HORCNN \cite{chao2018learning} as this branch is object invariant by design. They create a type of zero-shot learning dataset by reorganizing examples in the training and test sets of HICO-DET \cite{chao2018learning}, and using parts of the UnRel dataset \cite{peyre2017weakly} as a validation set. They do show great performance on zero-shot interaction categories, however we cannot rank their approach in Table \ref{tab:table1} as they do not rank their improvements against other models on HICO-DET. They propose their learning framework as an add-on to existing models.

We show the mAP scores on the HICO-DET dataset for most of the key models covered in this section in Table \ref{tab:table1}. The scores listed were found by their authors and published in their papers. HICO-DET offers several evaluation setups and difficulties shown in this table. Table \ref{tab:table2} contains scores for different models evaluated by average precision for a specified role in the V-COCO dataset.

\begin{table}
  \centering
  \begin{tabular}{l r r r r r r}
    & & \multicolumn{2}{c}{\small{\textbf{Default}}} & \multicolumn{3}{r}{\small{\textbf{Known Object}}} \\
    \cmidrule(r){2-4}
    \cmidrule(r){5-7}
    {\footnotesize \textit{Model}}
    & {\footnotesize \textit{Full}}
      & {\footnotesize \textit{Rare}}
      & {\footnotesize \textit{Non-Rare}}
      & {\footnotesize \textit{Full}}
      & {\footnotesize \textit{Rare}}
      & {\footnotesize \textit{Non-Rare}} \\
    \midrule
    \footnotesize HORCNN \cite{chao2018learning}  & 7.81 & 5.37 & 8.54 & 10.41 & 8.94 & 10.85 \\
    \footnotesize InteractNet \cite{gkioxari2018detecting}  & 9.94 & 7.16 & 10.77 & - & - & - \\
    \footnotesize GPNN \cite{Qi_2018_ECCV} & 13.11 & 9.34 & 14.23 & - & - & - \\
    \footnotesize iCAN\cite{gao2018ican}  & 14.84 & 10.45 & 16.15 & 16.26 & 11.33 & 17.73\\
    \footnotesize TIN \cite{li2018transferable} & 17.03 & 13.42 & 18.11 & 19.17 & 15.51 & 20.26 \\
    \footnotesize PMFNet \cite{wan2019pose} & 17.46 & 15.65 & 18.00 & 20.34 & 17.47 & 21.20\\
    \footnotesize RPNN \cite{zhou2019relation} & 17.35 & 12.78 & 18.71 & - & - & - \\
    \footnotesize VS-GAT\cite{liang2020visual}  & 20.27 & 16.03 & 21.54 & - & - & -\\
    \footnotesize PPDM \cite{liao2020ppdm} & 21.73 & 13.78 & 24.10 & 24.58 & 16.65 & 26.84 \\
    \footnotesize VCL \cite{hou2020visual} & 23.63 & 17.21 & 25.55 & 25.98 & 19.12 & 28.03 \\

  \end{tabular}
  \caption{Performance of the surveyed models on HICO-DET. dashes denote un-evaluated metrics from the original work (\%mAP)}~\label{tab:table1}
\end{table}

\begin{table}
  \centering
  \begin{tabular}{l r}
    {\small\textit{Model}}
    & {\small \textit{$AP_{role}$}}  \\
    \midrule
    Baseline \cite{gupta2015visual}  &  31.8\\
    InteractNet \cite{gkioxari2018detecting}  &  40.0\\
    GPNN \cite{Qi_2018_ECCV} & 44.0\\
    iCAN \cite{gao2018ican}  & 45.3\\
    iHOI \cite{xu2019interact} & 45.9\\
    RPNN \cite{zhou2019relation} & 47.5 \\
    VCL \cite{hou2020visual} & 48.3 \\
    TIN \cite{li2018transferable} & 48.7\\
    Cascaded HOI \cite{zhou2020cascaded} & 48.9 \\
    VS-GAT\cite{liang2020visual}  & 49.8\\
    PMFNet \cite{wan2019pose}  & 52.0\\
    
  \end{tabular}
  \caption{Performance of the surveyed models on V-COCO. (\%AP)}~\label{tab:table2}
\end{table}

\subsection{Datasets and Evaluation Metrics}

This section introduces the most common datasets used in the task of human-object interaction detection and provides insights on how they differ. High-quality datasets commonly contain localization and class labels on each of the objects or humans in the image. Human-object interaction detection requires image data to be labeled not only for objects but also for the relationships between the human and objects. For images with many instances of an interaction, these all must be separately labeled. Human-object interaction datasets must contain enough training data for all object classes as well as all relationship classes. Data for all possible real-world combinations of objects and relationships are impossible to obtain, therefore datasets typically pick a number of objects and interactions to focus on. There are many datasets used for this task, however, each dataset uses specific methods of providing ground truths, as well as different object and interaction classes. Each dataset also provides its own method of evaluating model performance. Table \ref{tab:table3} summarizes the datasets and their properties, as discussed in this section. 

One of the first purpose-built datasets for the task of human-object interaction detection is the Humans Interacting with Common Objects (HICO) \cite{chao2015hico} dataset, created by Chao et al. This dataset was constructed from the MS-COCO \cite{lin2014microsoft} dataset commonly used for object detection evaluation. HICO uses 80 object categories from MS-COCO and commonly used verbs to create the interaction categories for each object. Each object is also given a "no interaction" action, for a total of 600 human-object interactions. Each human-object interaction category has at a minimum of six images, and the test set should contain at least one image for that category. HICO does not provide instance level groundtruth annotations for every HOI occurring in each image. Another problem is the fact that images with multiple humans present are not exhaustively labeled. For example, in the case of a person riding in an airplane, there could be many people seated on board an airplane in the image, yet the HICO dataset would only require detecting a single HOI that fits that description. That is to say, that the HICO dataset proves image level groundtruth annotations. With these issues in mind Chen et al., the same authors of the HICO dataset, augment HICO to create HICO with Detection (HICO-DET) \cite{chao2018learning}. HICO-DET contains groundtruth labels for every human, and object participating in an annotated interaction class. The authors took the original HICO dataset and augmented it by crowd-sourcing the instance level groundtruth labeling via Amazon Mechanical Turk. 

The verbs in COCO (V-COCO) dataset \cite{gupta2015visual}, is another commonly evaluated dataset for human-object interaction detection. Similar to HICO, the object classes are taken from the COCO \cite{lin2014microsoft} dataset. But unlike HICO, the authors use the images already found in the COCO dataset. COCO has human-labeled and verified captions on each image, these are where the interaction classes are derived from. Using a simplified vocabulary, they designate 26 common actions amongst the different object classes. The COCO dataset contains ground truth labels for each object and human in the image, and the authors of V-COCO were able to reuse these. Another dataset, although less commonly used, for human-object interaction detection is the HCVRD dataset created by Zhuang et al. \cite{zhuang2018hcvrd}. This dataset is far more diverse in terms of labeled interactions and objects than the previously covered datasets. The images for HCVRD were gathered from the Visual Genome dataset \cite{krishna2017visual}, which contains object labels and bounding boxes, image captions, and labeled relationships between objects. The interactions included in HCVRD were drawn from the VG dataset where one of the objects is labeled as human. The authors took special care in "cleaning" the interactions by removing ambiguous actions and combining interactions with close similarity as a single interaction class. 

In human-object interaction detection, mean average precision (mAP) is most commonly used as an evaluation metric. For each image, the model should output a classification score for each interaction class. For each class, average precision is calculated from the entire test set of images. The mAP is computed as the average of the average precision scores. 
The authors provide an easy setting for evaluation called the "Known Object" setting. In this setting the verified positive images are used as positives with the verified negative images used as the negatives, skipping both the unknown and ambiguous images \cite{chao2015hico}. This removes the uncertainty of an imperfect object detector, by removing the images without the subject from the human-object interaction in question. For a more realistic setting, the authors propose adding the unknown category of images back as extra negatives. Testing on the HICO and HICO-DET datasets is done on both the Known Object setting as well as the realistic setting. Two common metrics for evaluation of models on the V-COCO dataset are agent detection and role detection \cite{gupta2015visual}. For agent detection, the task is to detect the humans performing a queried action. Average precision is used in this task as a performance metric, where humans labeled with the correct interaction category are marked positives. For role detection, the goal is to detect the human and objects participating in the given interaction. Models trained on HCVRD are tested against three metrics: predicate recognition where the interaction is detected given the bounding boxes for the human and object [50]. Phrase detection in which, given the human and object bounding boxes, the interaction as well as a union bounding box that encompasses the entire interaction or activity is predicted. For the final test metric relationship detection, measured in terms of recall, the model must localize the human and objects, as well as perform phrase detection. 

One last dataset to mention is the UnRel dataset \cite{peyre2017weakly}. UnRel is specifically created to evaluate unrealistic relationships between objects and people. However it specifically focuses on spatial relationships such as person-ride-dog or elephant-on-top-of-car, and includes non-human-object interactions. It can be used for add-on module training or in the case of \cite{song2020novel} where they manually filter out interaction classes that do not pertain to humans, as supplemental data. It is worth mentioning that a dataset of unrealistic interactions could help benefit future zero-shot and weakly supervised learning approaches to human-object interaction detection. 

\begin{table}
  \centering
  \begin{tabular}{l r r r}
    
    \cmidrule(r){2-4}
    {\small\textit{}}
    & {\small \textit{Images}}
      & {\small \textit{Interaction Classes}}
      & {\small \textit{Object Classes}} \\
    \midrule
    HICO \cite{chao2015hico} & 47,774 & 600 & 80 \\
    HICO-DET \cite{chao2018learning} & 47,776 & 600 & 80 \\
    V-COCO \cite{gupta2015visual} & 10,346 & 26 & 80 \\
    HCVRD \cite{zhuang2018hcvrd}  & 52,855 & 927 & 1824 \\
    
  \end{tabular}
  \caption{human-object interaction Dataset Metrics.}~\label{tab:table3}
\end{table}

\section{Proposed Work}
To understand and study the multi-stream architectures, we propose to implement the HORCNN detection model using the PyTorch deep learning framework. Following the implementation details, we re-created the model, which was originally built using Caffe. Other than the change in framework, a few deviations should be noted. The original authors use individual RCNN \cite{girshick2014rich} detectors for each object for their object detectors. We chose to use Faster-RCNN \cite{ren2015faster} to its immediate availability as a module included with PyTorch. In this implementation, Faster-RCNN is pre-trained on the MS-COCO dataset which includes the same object classes as HICO-DET. Another deviation is the use of AlexNet \cite{krizhevsky2012imagenet} rather than CaffeNet \cite{jia2014caffe}. For their implementation of HORCNN, the authors use CaffeNet pre-trained on the ImageNet dataset for the human and object convolutional streams. To avoid having to perform costly ImageNet pretraining, we used the pre-trained AlexNet implementation provided by Torchvision, widening the output feature vector to 600 classes to match the output classes of the HICO-DET dataset. AlexNet and CaffeNet are the same architecture, with AlexNet being modified for multi GPU use. With our implemented version of the HORCNN model, we will follow the training process as detailed in the original work. With this trained model, we will perform an extensive ablation study where we break the model into its various streams and combinations. With the results found in this study we should be able to see which model components and which image features give the best performance for to HICO-DET dataset. Our prediction is that the human centric component of the network should provide the model the most detailed visual features for predicting a human-object interaction class. This is driven by the understanding that a human should exhibit the intent to interact on the object. For example, features extracted from a human kicking a sports ball should be far different than features extracted from a human throwing a sports ball. From the other end, we should see that the features extracted from a sports ball being kicked should remain similar to those of a sports ball being thrown. 

\subsection{Experiments and Results}

For the re-implementation of the HORCNN model, we see a close but slightly reduced mAP on the HICO-DET dataset, close to that of the original papers. Differences could be explained by hyperparameter adjustments. Due to computational constraints, our implementation was trained with a batch size of four images, containing four randomly sampled proposals from the true positive, type I negative, and type II negative proposal sets, listed in the previous section of this paper. Using the batch sizes results in a total batch size of 16 proposals. In the original work, eight images are selected per batch, with 8 proposals per image, for an overall batch size of 64 proposals. We trained four times as long as the original work due to the reduction in proposals from our training parameters. We trained for 400k iterations at a learning rate of 0.001, and 200k iterations at a learning rate of 0.0001. Results and comparisons can be seen in Table \ref{tab:table4} and Table \ref{tab:table5}. The model was trained for ~20 hours on a single Nvidia TitanXp GPU.

\begin{table}
  \centering
  \begin{tabular}{l r r r}
    \cmidrule(r){2-4}
    {\small\textit{Implementation}}
    & {\small \textit{Full}}
      & {\small \textit{Rare}}
      & {\small \textit{Non-Rare}} \\
    \midrule
    Ours & 5.87	& 3.06 & 7.08 \\
    Chao et al. \cite{chao2018learning}& 7.81 & 5.37 & 8.54 \\
  \end{tabular}
  \caption{Performance (\%mAP) of re-implemented model vs. published results}~\label{tab:table4}
\end{table}

The three streams of the HORCNN model, human, object, and pairwise streams, extract fine-grained features from their subjects. However, these feature weights are summed when making a final prediction on whether a human-object pair is engaged in an interaction. From a general understanding of human interaction, we know that fine-grained features such as body placement and pose can influence a decision on whether or not a human is interacting with an object. We perform studies on each individual stream and selected combinations to see if one performs best in the overall task of identifying an interaction. Results of these tests can be seen in Table \ref{tab:table5}. We evaluated several test cases for mAP score. The evaluation was done over the entire testing set, using 10 proposals from each image, similar to how the authors of HORCNN perform their evaluations. HOP denotes the full model including scores from the human, object, and pairwise streams. H, O, and P denote human, object, and pairwise streams respectively. HO denotes the score of the human and object branches combined. Finally, HP denotes the human and pairwise streams combined. 

\begin{table}
  \centering
  \begin{tabular}{l r r r}
    \cmidrule(r){2-4}
    {\small\textit{Implementation}}
    & {\small \textit{Full}}
      & {\small \textit{Rare}}
      & {\small \textit{Non-Rare}} \\
    \midrule
    Full Model & 5.87 & 3.06 & 7.08 \\
    H & 1.62 & 0.40 & 2.09 \\
    O & 4.65 & 2.78 & 5.67 \\
    P & 0.93 & 0.07 & 1.08 \\
    H, O & 5.41 & 3.51 & 6.54 \\
    H, P & 1.41 & 0.15 & 1.65 \\

  \end{tabular}
  \caption{Performance of individual model streams (\%mAP). H, O,P denote Human, Object, and Pairwise streams respectively }~\label{tab:table5}
\end{table}

Our original hypothesis was that the human stream would be more dominant in guiding predictions, however, the results show that the object stream has the best mAP on the test set and seems to be the dominant factor in the HORCNN model. We believe that this is caused by similar interactions between multiple object categories. For example, the interaction ‘carry’ is valid for 32 of the 80 object categories. While many of the human appearances could be similar for certain groupings of objects, it is likely that there is not enough information from the human appearances alone to differentiate between these exact object interaction classes. This can be seen in some of the results from test images on the trained model, where similar interactions between objects receive relatively high scores. When combining the human and the object streams, we see that the mAP improves slightly over just the object stream, however, it performs better against the combination of the human and pairwise streams. This shows the importance of the object stream in making predictions on the HOI classes. Unsurprisingly, the full model incorporating all the streams achieves the highest mAP score, this proves the importance of incorporating all three streams in the HORCNN model. Out of the previous works surveyed in the related works section of this paper, HORCNN achieves the lowest mAP scores, quite low for a good prediction model.

\subsection{Dataset Discussion}

\begin{figure}
\centering
  \includegraphics[width=0.7\columnwidth]{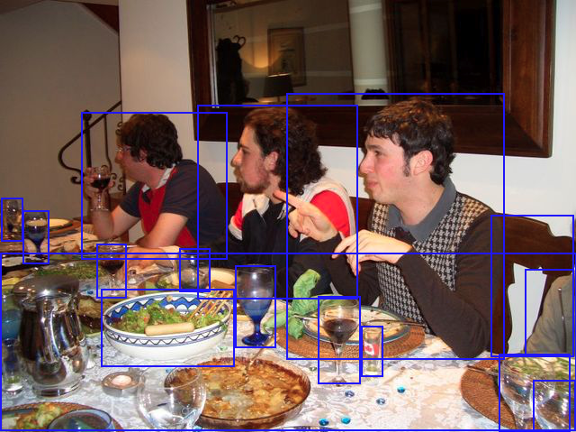}
  \caption{Example of non-exhaustively labeled image from HICO-DET }~\label{fig:figure4}
\end{figure}

The HICO-DET dataset is large and fairly diverse, however, there are a few issues present. First, for each object category, there is a ‘no interaction’ class. This provides samples for a model to learn how to distinguish when there are objects and humans in an image, but they are not interacting with each other. However, there are many instances in the images where there should be a no interaction category, but they are not labeled as participating in an interaction with a human. While training with the image centric sampling strategy, the model could be given these samples, since samples are chosen at random, without a label and will be penalized in the loss function since no ground truth exists. It is possible to hand label these human-object proposals with a ‘no interaction’ proposal while loading the data, but in doing so the dataset becomes imbalanced. Interestingly, there are some human-object pairs that are participating in an interaction class in images in the dataset, that are not labeled. For example, the image in Figure \ref{fig:figure4} is taken from the HICO-DET training set with bounding boxes representing detections from FastRCNN. The ground truth annotations only contain labels for four separate humans ‘sit at’ and ‘eat at’ dining table. But clearly, we see that one human is drinking from and holding a cup, as well as many cups and plates in the image that should be labeled with ‘no interaction’.  

The fact that human to human relationships are present in the HICO-DET dataset, provides an extra complexity when searching for proposals. Unfortunately, all images containing multiple humans does not have a ‘no interaction’ label between these human detections. Since the object detection selection must pair all humans with all objects, and all humans with all humans, it is likely that one of these unlabeled human to human relationships show up in the dataset. While it is possible to create these labels artificially in the data loader, it adds more unnecessary data preprocessing for the training. And it is not guaranteed that these labels are true ‘no interaction’ labels, instead of missed interactions. 

\begin{figure}
\centering
  \includegraphics[width=0.7\columnwidth]{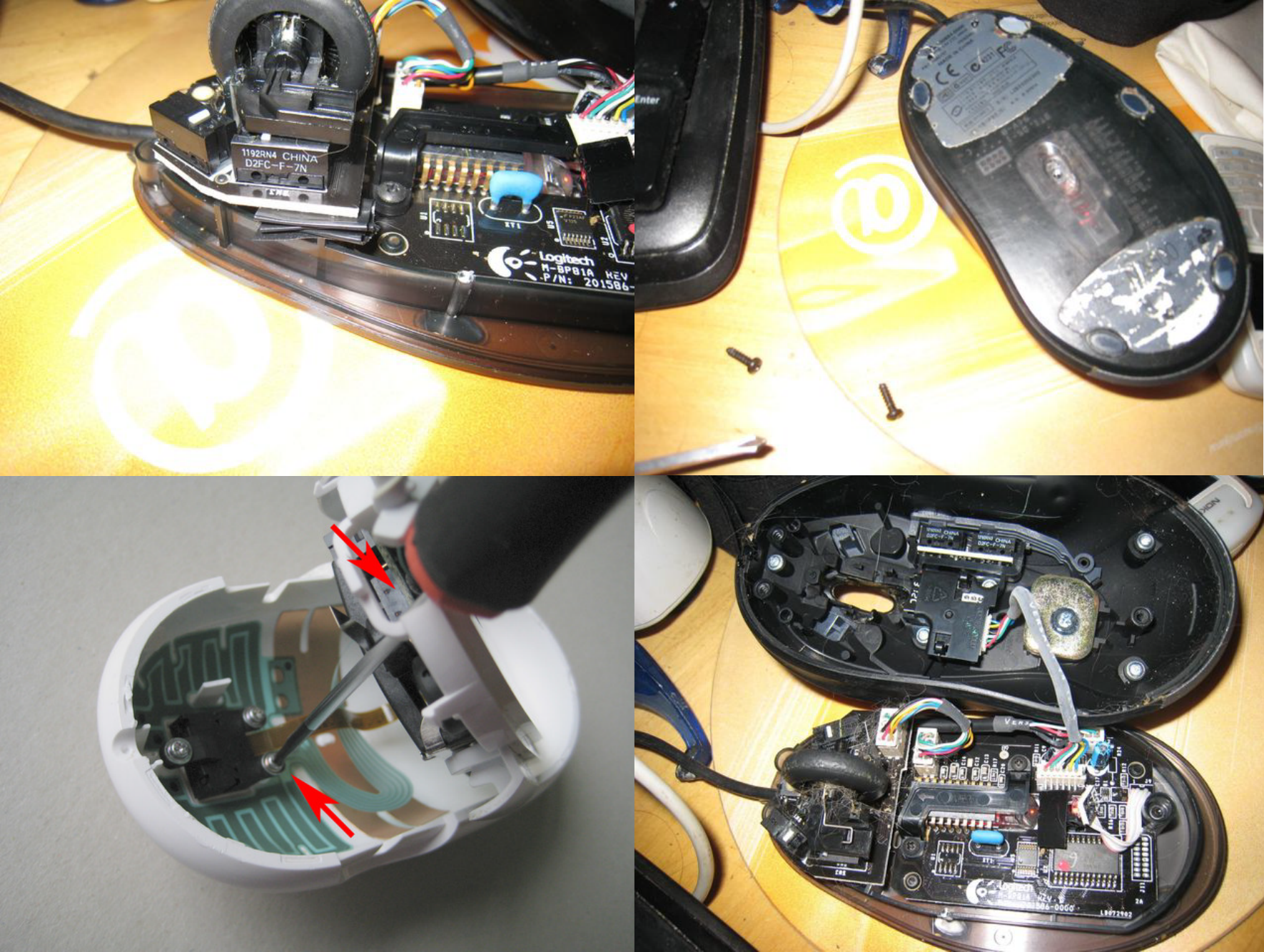}
  \caption{Examples of the interaction class ‘human repair mouse’ from the HICO-DET dataset. }~\label{fig:figure5}
\end{figure}

HICO-DET contains a number of rare human-object interaction classes, as evidenced by the ‘rare’ setting for evaluation. However, the quality of these examples leaves doubt in the ability of the human reviewers to filter out poor images, or images that do not display the interaction. For example, the image seen in Figure \ref{fig:figure5} contains training and test images labeled as containing the relationship of human-repair-mouse, mouse in this context referring to a computer mouse. It is clear from this picture that there is no human present in the image. An automated data-processing pipeline would not label this as the interaction class human-repair-mouse, these are the only training examples for this interaction in the entire dataset. This issue could be present in other small objects in the dataset; however, we find this to be the most egregious error. This brings into question the quality of the HICO-DET dataset, and its ability to train high performing models for human-object interaction detection. 

\subsection{Conclusion}
In this work we have taken an in-depth examination of the task of human-object interaction detection, covering datasets and the baseline models. We performed studies on the baseline model for the HICO-DET dataset, HORCNN, to identify the most robust model components and features. We see that for the multi stream approach presented in HORCNN, the object appearance features provide the most accurate prediction on the dataset. However, it does not compare to the combination of the streams to provide accurate human-object interaction detections. We hope that the findings of these studies can influence future model design in this field of research. 

The HICO-DET dataset for human-object interaction detection was also examined throughout this work. We have shown some concerning quality issues regarding this dataset. It is our opinion that this dataset should be more carefully examined for accurate labeling and higher quality images, especially for the crucial training segment of the dataset. With some updating, this dataset could become very valuable to researchers in this field.


\bibliographystyle{ACM-Reference-Format}
\bibliography{ACM_MM_2020_HOIDET_SURVEY}

\appendix

\end{document}